%% file: paper2691.tex
\documentclass[runningheads]{llncs}

\usepackage{graphicx}
\usepackage[hidelinks]{hyperref}
\usepackage{booktabs} 
\usepackage{tabularx} 
\usepackage{colortbl}
\usepackage{multirow}
\usepackage{amsfonts} 
\usepackage{todonotes}
\usepackage{caption} 
\usepackage{chngpage}
\usepackage{xcolor}
\newcommand{\oldtext}[1]{\textcolor{black}{#1}}
\newcommand{\newtext}[1]{\textcolor{black}{#1}} 
\newcommand{\rephrasedtext}[1]{\textcolor{black}{#1}} 
  
\begin{document}
\input{1_title}
\input{2_abs}
\input{3_intro}
\input{4_related}
\input{5_methods}

\input{6_exps}
\input{7_discuss}
\input{8_bib}
\end{document}

%% file: 1_title.tex
\title{Boundary-weighted logit consistency improves calibration of segmentation networks}
%
%
\titlerunning{Boundary-weighted logit consistency improves segmentation calibration}
%
\author{Neerav Karani \and Neel Dey \and Polina Golland}

\institute{Massachusetts Institute of Technology}

\authorrunning{Karani et al}
%
\maketitle              

%% file: 2_abs.tex
\begin{abstract}
\newtext{Neural network prediction probabilities and accuracy are often only weakly-correlated.}
\rephrasedtext{Inherent label ambiguity in training data for image segmentation aggravates such miscalibration.}
We show that logit consistency across stochastic transformations acts as a spatially varying regularizer that prevents
\rephrasedtext{overconfident}
predictions at pixels with ambiguous labels.
Our boundary-weighted extension of this
\rephrasedtext{regularizer}
provides state-of-the-art calibration for
\rephrasedtext{prostate and heart MRI segmentation}.
\end{abstract}

%% file: 3_intro.tex
\section{Introduction}~\label{sec_intro}
Supervised learning of deep neural networks is susceptible to overfitting when labelled training datasets are small, as is often the case in medical image analysis.
Data augmentation (DA) tackles this issue by
\rephrasedtext{transforming}
(informed by knowledge of task-specific invariances and equivariances)
labelled input-output pairs, thus simulating
\rephrasedtext{new}
input-output pairs to expand the training dataset.
\rephrasedtext{This idea is used in semi-supervised learning \cite{bortsova2019semi,li2020transformation} via}
an unsupervised loss function that promotes the desired invariance and equivariance properties in
predictions for unlabelled images.
We refer to this as consistency regularization (CR).

\vspace{4pt} \noindent 
While previous work has employed CR to leverage unlabelled images, we show that \emph{even in the absence of any additional unlabelled images}, CR improves \emph{calibration} \cite{guo2017calibration}, and sometimes, even segmentation \emph{accuracy} of neural networks over those trained with DA.
This is surprising at first sight.
Compared to DA, when employed in the supervised setting, CR does not have access to additional data.
What are then the causes of this benefit?

\vspace{4pt} \noindent 
To answer this question, we note that boundaries between anatomical regions are often ambiguous in medical images due to absence of sufficient contrast or presence of image noise or partial volume effects.
Annotations in labelled segmentation datasets, however, typically comprise of \textit{hard} class assignments for each pixel, devoid of information regarding such ambiguity.
Supervised learning approaches then insist on \emph{perfect agreement} at every pixel between predictions and ground truth labels, which can be achieved by over-parameterized neural networks.
For instance, using the cross-entropy loss function for training maximizes logit differences between the ground truth class and other classes for each pixel \cite{murugesan2022calibrating}.
This bias for low-entropy predictions caused by supervised learning loss functions coupled with inherent ambiguity in the true underlying labels leads to over-confident predictions and miscalibrated models.

\vspace{4pt} \noindent
This viewpoint suggests that reduced logit differences across classes for pixels with ambiguous labels may help counter such miscalibration.
Based on this idea, we make two main contributions in this paper.
First, we show that CR can automatically discover such pixels and prevent overfitting to their \emph{noisy} labels.
In doing so, CR induces a spatially varying pixel-wise regularization effect, leading to improved calibration.
In contrast to previous use of CR in medical image segmentation \cite{bortsova2019semi,li2020transformation}, these new benefits are independent of additional unlabelled images.
Second, based on this understanding of the mechanism underlying the calibration benefits of CR, we propose a spatially-varying weighing strategy for the CR loss relative to the supervised loss.
This strategy emphasizes regularization in pixels near tissue boundaries, as these pixels are more likely to suffer from label ambiguity.
We illustrate the calibration benefits of our approach on segmentation tasks in prostate and heart MRI.

%% file: 4_related.tex
\section{Related Work}~\label{sec_related}
\textbf{Label ambiguity in medical image segmentation} is tackled either by generating multiple plausible segmentations for each image \cite{baumgartner2019phiseg,kohl2018probabilistic}, or by predicting a single well-calibrated segmentation \cite{hong2021hypernet,islam2021spatially,larrazabal2021maximum,mehrtash2020confidence,murugesan2022calibrating}.
In the latter group, predictions of multiple models are averaged to produce the final segmentation \cite{hong2021hypernet,mehrtash2020confidence}.
Alternatively, the training loss of a single model is modified to prevent low-entropy predictions at all pixels \cite{murugesan2022calibrating}, at pixels with high errors \cite{larrazabal2021maximum} or pixels near boundaries \cite{islam2021spatially,murugesan2023trust}.
Smoothing ground truth labels of boundary pixels \cite{islam2021spatially} disregards image intensities that cause label ambiguity.
In contrast, the boundary-weighted variant of our approach emphasizes regularization in those regions but allows consistency across stochastic transformations to differentiate sub-regions with varying label ambiguity.
Related, boundary-weighted supervised losses have been proposed in different contexts \cite{abulnaga2022automatic,hoopes2022synthstrip}.

\vspace{4pt} \noindent \textbf{Aleatoric uncertainty estimation} in medical images \cite{monteiro2020stochastic,wang2019aleatoric}
is closely related to the problem of pixel-wise label ambiguity due to uncertainty in the underlying image intensities.
In particular, employing stochastic transformations during inference has been shown to produce estimates of aleatoric uncertainty \cite{wang2019aleatoric}, while we use them during training to automatically identify regions with ambiguous labels and prevent low-entropy segmentation predictions in such regions.

\vspace{4pt} \noindent In \textbf{semi-supervised medical image segmentation}, CR is widely used as a means to leverage unlabelled images to improve segmentation accuracy \cite{bortsova2019semi,cui2019semi,li2020transformation,luo2022semi,wu2022mutual}.
In contrast, we investigate the capability of CR to improve calibration without using any unlabelled images.
Finally, for \textbf{image classification}, CR can help mitigating label noise \cite{englesson2021generalized} and label smoothing has been shown to improve calibration \cite{muller2019does}.
To our knowledge, this paper is the first to investigate the role of CR as a means to improve calibration of segmentation models.

%% file: 5_methods.tex
\section{Methods}~\label{sec_methods}
\input{5_methods_part1}
\input{5_methods_part2}
\input{5_methods_part3}

%% file: 5_methods_part1.tex
Using a labelled dataset $\{(X_i, Y_i)\}, i=1,2, \dots n$, we wish to learn a function that maps images $X \in \mathbb{R}^{H \times W}$ to segmentation labelmaps $Y \in \{1, 2, ..., C\}^{H \times W}$, where $C$ is the number of classes.
Let $f_\theta$ be a convolutional neural network that predicts $\hat{Y} = \sigma(f_\theta(X))$, where $f_\theta(X) \in \mathbb{R}^{H \times W \times C}$ are logits and $\sigma$ is the softmax function.
In supervised learning, optimal parameter values are obtained by minimizing an appropriate supervised loss, $\hat{\theta} = \arg\!\min_{\theta} \: \mathbb{E}_{X,Y} \: \mathcal{L}_{\textrm{s}}(\sigma(f_\theta(X)), Y)$.

\vspace{4pt} \noindent \textbf{Data augmentation (DA)}
leverages knowledge that the segmentation function is invariant to intensity transformations $S_\phi$ (e.g., contrast and brightness modifications, blurring, sharpening, Gaussian noise addition) and equivariant to geometric transformations $T_\psi$ (e.g., affine and elastic deformations).
The optimization becomes
$\hat{\theta} = \arg\!\min_{\theta} \: \mathbb{E}_{X,Y,\phi, \psi} \: \mathcal{L}_{\textrm{s}}(\sigma(g(X;\theta,\phi,\psi)), Y)$, where $g(X;\theta,\phi,\psi) = T_\psi^{-1}(f_\theta(S_\phi(T_\psi(X))))$.
In order to achieve equivariance with respect to $T_\psi$, the loss is computed after applying the inverse transformation to the logits.


\vspace{4pt} \noindent \textbf{Consistency regularization (CR)}
additionally constrains the logits predicted for similar images to be similar.
This is achieved by minimizing a consistency loss $\mathcal{L}_{\textrm{c}}$ between logits predicted for two transformed versions of the same image:
$\hat{\theta} = \arg\!\min_{\theta} \: \mathbb{E}_{X,Y,\phi,\psi,\phi',\psi'} \: \mathcal{L}_{\textrm{s}}(\sigma(g(X;\theta,\phi,\psi)), Y) \: + \: \lambda \: \mathcal{L}_{\textrm{c}}(g(X;\theta,\phi,\psi), g(X;\theta,\phi',\psi'))$.
The exact strategy for choosing arguments of $\mathcal{L}_{\textrm{c}}$ can vary: as above, we use predictions of the same network $\theta$ for different transformations ($\phi$, $\psi$) and ($\phi'$, $\psi'$) \cite{bortsova2019semi};
alternatives include setting $\phi' = \phi$, $\psi' = \psi$, and using two variants of the model $\theta$ and $\theta'$ \cite{sajjadi2016regularization,tarvainen2017mean} or different combinations of these approaches \cite{cui2019semi,li2020transformation}.

%% file: 5_methods_part2.tex
\subsection{Consistency regularization at pixel-level}~\label{sec_cr_pixel}
Here, we show how understanding the relative behaviours of the supervised and unsupervised losses used in CR help to improve calibration.
Common choices for $\mathcal{L}_{\textrm{s}}$ and $\mathcal{L}_{\textrm{c}}$ are pixel-wise cross-entropy loss and pixel-wise sum-of-squares loss, respectively. For these choices, the total loss for pixel $j$ can be written as follows:

\begin{equation}~\label{eqn_pixel_wise}
    \mathcal{L}^j = \mathcal{L}_{\textrm{s}}^j + \lambda \: \mathcal{L}_{\textrm{c}}^j
    \\ = - \: \sum_{c=1}^C y_{c}^{j} \: \log (\sigma(z_{c}^{j})) + \lambda \: \sum_{c=1}^C (z_{c}^{j} - z_{c}^{\prime j})^{2} \textrm{,}
\end{equation}

\noindent where $z^{j}$ and $z^{\prime j}$ are $C$-dimensional logit vectors at pixel $j$ in $g(X;\theta,\phi,\psi)$ and $g(X;\theta,\phi',\psi')$ respectively, and the subscript $c$ indexes classes.

\vspace{4pt} \noindent $\mathcal{L}_{\textrm{s}}$ drives the predicted probability of the ground truth label class to $1$, and those of all other classes to $0$. Such low-entropy predictions are preferred by the loss function even for pixels whose predictions should be ambiguous due to insufficient image contrast, partial volume effect or annotator mistakes.

\vspace{4pt} \noindent Consistency loss $\mathcal{L}_{\textrm{c}}$ encourages solutions with consistent logit predictions across stochastic transformations. This includes, but is not restricted to, the low-entropy solution preferred by $\mathcal{L}_{\textrm{s}}$.
In fact, it turns out that due to the chosen formulation of $\mathcal{L}_{\textrm{s}}$ in the probability space and $\mathcal{L}_{\textrm{c}}$ in the logit space, deviations from logit consistency are penalized more strongly than deviations from low-entropy predictions. Thus, $\mathcal{L}_{\textrm{c}}$ permits high confidence predictions only for pixels where logit consistency across stochastic transformations can be achieved.

\vspace{4pt} \noindent Furthermore, variability in predictions across stochastic transformations has been shown to be indicative of aleatoric image uncertainty \cite{wang2019aleatoric}. This suggests that inconsistencies in logit predictions are likely to occur at pixels with high label ambiguity, causing high values of $\mathcal{L}_{\textrm{c}}$ and preventing high confidence predictions at pixels with latent ambiguity in labels.

\begin{figure}[t]
    \centering
    \includegraphics[width=0.99\textwidth, trim={0cm 10.1cm 0cm 0cm},clip]{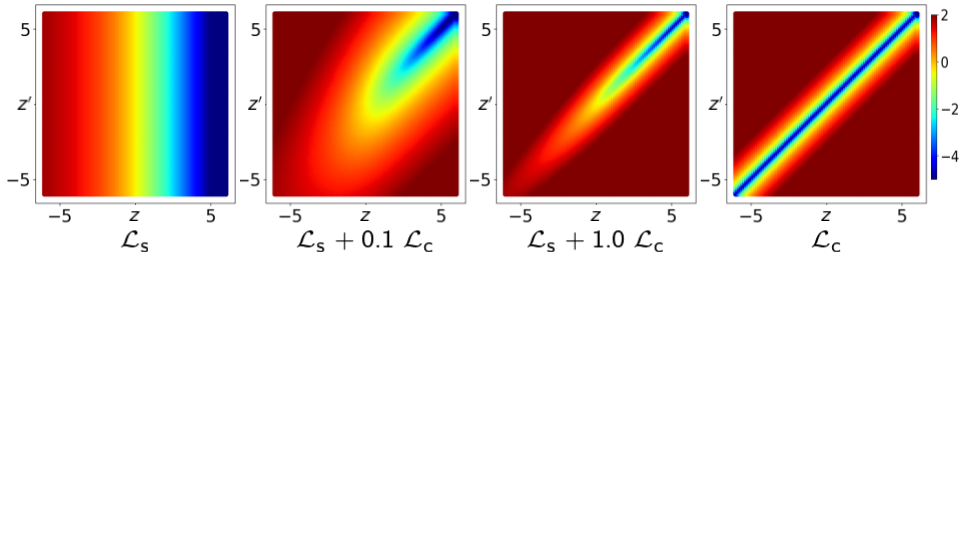} 
    \caption{Loss landscapes shown in log scale for $\mathcal{L}_{\textrm{s}}$ (left), $\mathcal{L}_{\textrm{c}}$ (right) and the total loss in Eqn. \ref{eqn_pixel_wise} for different values of $\lambda$ (center two), as $z$ and $z'$ vary.}
    \label{fig_binary_loss_landscape}
\vspace{-10pt}
\end{figure}

\vspace{4pt} \noindent \textbf{Special case of binary segmentation}:
To illustrate the pixel-wise regularization effect more clearly, let us consider binary segmentation.
\rephrasedtext{Here,}
we can fix $z_1 = 0$ and let $z_2 = z$, as only logit differences
\rephrasedtext{matter}
in the softmax function. Further, let us consider only one pixel, drop the pixel index and assume that its
\oldtext{ground truth}
label is $c = 2$. Thus, $y_1 = 0$ and $y_2 = 1$. With these simplifications, $\mathcal{L_\textrm{s}} = - \: \log (\sigma(z))$ and $\mathcal{L_\textrm{c}} =  (z - z')^2$.
Fig.~\ref{fig_binary_loss_landscape} shows that $\mathcal{L}_{\textrm{s}}$ favours high $z$ values, regardless of $z'$, while $\mathcal{L}_{\textrm{c}}$ prefers
the $z = z'$ line, and heavily penalizes deviations from it.
The behaviour of these losses is similar for multi-label segmentation.

%% file: 5_methods_part3.tex
\subsection{Spatially varying weight for consistency regularization}~\label{sec_bwcr}
Understanding consistency regularization as mitigation against overfitting to
\rephrasedtext{hard}
labels in ambiguous pixels points to a straightforward improvement of
\rephrasedtext{the}
method.
Specifically, the regularization term in the overall loss should be weighed higher when higher pixel ambiguity, and thus, higher label noise, is expected.
Natural candidates for higher ambiguity are pixels near label boundaries.
Accordingly, we propose boundary-weighted consistency regularization (BWCR):

\begin{equation}~\label{eqn_pixel_wise_boundary_weighted}
    \mathcal{L}^j = \mathcal{L}_{\textrm{s}}^j + \lambda(r^j) \: \mathcal{L}_{\textrm{c}}^j
\end{equation}

\begin{equation}~\label{lambda_drop}
    \lambda(r^j) = \lambda_{\textrm{max}} \Big{(}\frac{max(R-r^j, 0)}{R}\Big{)} + \lambda_{\textrm{min}}
\end{equation}

\begin{figure}[t]
    \centering
    \includegraphics[width=0.99\textwidth, trim={0cm 13.2cm 0cm 0cm},clip]{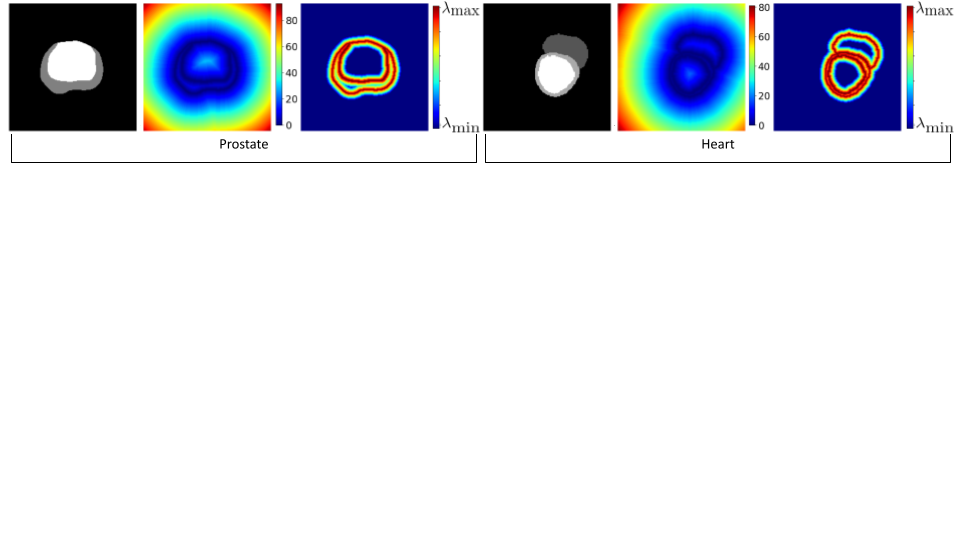}
    \caption{\newtext{From left to right, ground truth label, absolute distance function map, and proposed spatially varying weight for consistency regularization for $R = 10$.}}
    \label{fig_lambda_bwcr}
\vspace{-10pt}
\end{figure}

\noindent
\rephrasedtext{where $r^j$ is the distance to the closest boundary from pixel $j$},
$\lambda(r^j)$ drops away from the label boundaries, and $R$ is the width of the boundary region affected by the regularization.
\newtext{We compute $r^j = \textrm{argmin}_c \: r^j_c$, where $r^j_c$ is the absolute value of the euclidean distance transform \cite{paglieroni1992distance} at pixel $j$ of the binarized segmentation for foreground label $c$.
Fig.~\ref{fig_lambda_bwcr} shows examples of $r^j$ and $\lambda(r^j)$ maps.}

%% file: 6_exps.tex
\section{Experiments and Results}~\label{sec_exps}
\input{6_exps_part1}
\input{6_exps_part2}
\input{6_exps_part3}

%% file: 6_exps_part1.tex
\textbf{Datasets}: We investigate the effect of CR on two public datasets.
The NCI \cite{bloch2015nci}
dataset includes T2-weighted prostate MRI scans of $N = 70$ subjects
\rephrasedtext{(30 acquired with a 3T scanner and a surface coil, and 40 acquired with a 1.5T scanner and an endo-rectal coil).}
In-plane resolution is $0.4$ - $0.75$ mm$^2$, through-plane resolution is $3$ - $4$mm.
Expert annotations are available for central gland (CG) and peripheral zone (PZ). 
The ACDC \cite{bernard2018deep}
dataset consists of cardiac cine MRI scans of $N = 150$ subjects
\oldtext{(evenly distributed over 4 pathological types and healthy subjects, and acquired using 1.5T and 3T scanners).}
In-plane resolution is $1.37$ - $1.68$ mm$^2$, through-plane resolution is $5$ - $10$ mm.
Expert annotations are provided for right ventricle (RV), left ventricle (LV) and myocardium (MY).
Two $3$D volumes that capture the end-systolic and end-diastolic stages of the cine acquisition respectively are annotated for each subject.

\vspace{4pt} \noindent \textbf{Data splits}:
From the $N$ subjects in each dataset, we select $N_{\textrm{ts}}$ test, $N_{\textrm{vl}}$ validation and $N_{\textrm{tr}}$ training subjects.
$\{N_{\textrm{ts}}, N_{\textrm{vl}}\}$ are set to $\{30, 4\}$ for NCI and $\{50, 5\}$ for ACDC.
We have $3$ settings for $N_{\textrm{tr}}$: \emph{small}, \emph{medium} and \emph{large},
with $N_{\textrm{tr}}$ as $6$, $12$ and $36$ for NCI, and $5$, $10$ and $95$ for ACDC, in the three settings, respectively.
All experiments are run thrice, with test subjects fixed across runs, and training and validation subjects randomly sampled from remaining subjects.
\oldtext{In each dataset, subjects in all subsets are evenly distributed over different scanners.}


\vspace{4pt} \noindent \textbf{Pre-processing}: We
correct bias fields using the N4 \cite{tustison2010n4itk} algorithm,
linearly rescale intensities of each image volume using its $2$nd and $98$th intensity percentile, followed by clipping at $0.0$ and $1.0$,
resample (linearly for images and with nearest-neighbours for labels) NCI and ACDC volumes to $0.625$ mm$^2$ and $1.33$ mm$^2$ in-plane resolution, while leaving the through-plane resolution unchanged, and crop or pad with zeros to set the in-plane size to $192$ x $192$ pixels.


\vspace{4pt} \noindent \textbf{Training details}: We use a $2$D U-net \cite{ronneberger2015u} architecture for $f_\theta$, and use cross-entropy loss as $\mathcal{L}_{\textrm{s}}$ and squared difference between logits as $\mathcal{L}_{\textrm{c}}$.
\rephrasedtext{For $S_\phi$, we employ gamma transformations, linear intensity scaling and shifts, blurring, sharpening and additive Gaussian noise. For $T_\psi$, we use affine transformations. For both, we use the same parameter ranges as in \cite{zhang2020generalizing}.}
For every $2$D image in a batch, we apply each transformation with probability $0.5$.
We set the batch size to $16$, train for $50000$ iterations with Adam optimizer, and linearly decay the learning rate from $10^{-4}$ to $10^{-7}$.
After the training is completed, we set $\theta$ to its exponential moving average at the iteration with the best validation Dice score \cite{arpit2021ensemble}.


\vspace{4pt} \noindent \textbf{Evaluation Criteria}: We evaluate segmentation accuracy using Dice similarity coefficient and calibration using Expected Calibration Error (ECE) \cite{guo2017calibration} and Thresholded Adaptive Calibration Error (TACE) \cite{nixon2019measuring} (computed using $15$ bins and threshold of $0.01$).
ECE measures the average difference of accuracy and mean confidence of binned predicted probabilities, while TACE employs an adaptive binning scheme such that all bins contain an equal number of predictions.



%% file: 6_exps_part2.tex
\subsection{Effect of CR}
\rephrasedtext{First, we check if CR improves calibration of segmentation models.}
We perform this experiment in the small training dataset setting, and present results in Table \ref{tab_cr_lambda}.
It can be seen that as $\lambda$ (Eqn. \ref{eqn_pixel_wise}) increases from $0.01$ to $1.0$, CR improves calibration in both datasets while retaining similar segmentation accuracy to DA ($\lambda = 0.0$).
These results validate the discussion presented in Sec. \ref{sec_cr_pixel}.
However, increasing $\lambda$ to $10.0$ leads to accuracy degradation.
This motivates us to propose the boundary-weighted extension to CR in order to further improve calibration while preserving or improving segmentation accuracy.

\begin{table}[h]
    \centering
    \begin{tabular}{l|ccc|ccc}
        \hline
        Method & \multicolumn{3}{c|}{NCI} & \multicolumn{3}{c}{ACDC} \\

        \hline
        $\lambda$ & Dice $\uparrow$ & ECE $\downarrow$ & TACE $\downarrow$ & Dice $\uparrow$ & ECE $\downarrow$ & TACE $\downarrow$ \\

        \hline

        $0.0$ &
        66$\pm$13 & 24$\pm$14 & 11$\pm$4 &
        76$\pm$12 & 20$\pm$14 & 10$\pm$4 \\
        
        $0.01$ &
        66$\pm$13 & 25$\pm$13 & 12$\pm$4 &
        75$\pm$13 & 19$\pm$14 & 9$\pm$3 \\

        $0.1$ &
        66$\pm$14 & 24$\pm$14 & 11$\pm$3 &
        75$\pm$12 & 17$\pm$14 & 7$\pm$3 \\

        $1.0$ &
        65$\pm$14 & 18$\pm$14 & 6$\pm$3 &
        75$\pm$12 & 13$\pm$12 & 5$\pm$2 \\

        $10.0$ &
        63$\pm$13 & 13$\pm$12 & 1$\pm$1 &
        70$\pm$12 & 16$\pm$11 & 1$\pm$0 \\

        \hline
        
    \end{tabular}
    
    \caption{Effect of CR ($\lambda > 0$) and DA ($\lambda = 0$) on segmentation accuracy and calibration.
    Results are reported as $\%$ average $\pm$ $\%$ standard deviation values of over test volumes and three experiment runs.
    For brevity, TACE values are scaled by $10$.
    Increasing $\lambda$ from $0.01$ to $1.0$ improves calibration, but further increasing $\lambda$ leads to degradation in segmentation accuracy.}
    \label{tab_cr_lambda}
\end{table}
\vspace{-35pt}

%% file: 6_exps_part3.tex
\subsection{Effect of BWCR}
We compare CR and BWCR with the following baseline methods:
(1) supervised learning without DA (Baseline),
(2) data augmentation (DA) \cite{zhang2020generalizing},
(3) spatially varying label smoothing (SVLS) \cite{islam2021spatially} and (4) margin-based label smoothing (MLS) \cite{liu2022devil,murugesan2022calibrating}.
For CR, we set $\lambda = 1.0$.
For BWCR, we set $\lambda_{\textrm{min}} = 0.01$, $\lambda_{\textrm{max}} = 1.0$ and $R = 10$ pixels.
\newtext{These values were set heuristically; performance may be further improved by tuning them using a validation set.}
For SVLS and MLS, we use the recommended hyper-parameters, setting the size of the blurring kernel to $3 \times 3$ and its standard deviation to $1.0$ in SVLS, and margin to $10.0$ and
\rephrasedtext{regularization term weight}
to $0.1$ in MLS.
To understand the behaviour of these methods under
\rephrasedtext{different training dataset sizes,}
we carry out these comparisons in the \emph{small}, \emph{medium} and \emph{large} settings explained above.
The following observations can be made from Table \ref{tab_quant_results}:

\begin{enumerate}
    \item As training data increases, both accuracy and calibration of the supervised learning baseline improve. Along this axis, reduced segmentation errors improve calibration metrics despite low-entropy predictions.
    
    \item Similar trends exist for DA along the data axis.
    For fixed training set size, DA improves both accuracy and calibration due to the same reasoning as above. This indicates that strong DA should be used as a baseline method when developing new calibration methods.
    
    \item Among the calibration methods, CR provides better calibration than SVLS and MLS. BWCR improves calibration even further.
    For all except the ACDC \emph{large} training size setting, BWCR's improvements in ECE and TACE over all other methods are statistically significant ($p < 0.001$) according to paired permutation tests.
    Further, while CR causes slight accuracy degradation compared to DA, BWCR improves or retains accuracy in most cases.
    \item Subject-wise calibration errors (Fig. \ref{fig_subjectwise_ECE}) show that improvements in calibration statistics stem from consistent improvements across all subjects.
    \item Fig. \ref{fig_qualitative_results} shows that
    \rephrasedtext{predictions of CR and BWCR are less confident}
    around boundaries.
    BWCR also
    \rephrasedtext{shows}
    different uncertainty in pixels with similar distance to object boundaries but different levels of image uncertainty.
    \newtext{\item Fig. \ref{fig_qualitative_results} also reveals an intriguing side-effect of the proposed method: CR, and to a lesser extent BWCR, exhibit \emph{confidence leakage} along object boundaries of other foreground classes. For instance, in row $1$ ($3$), CR assigns probability mass along PZ (MY) edges in the CG (RV) probability map. We defer analysis of this behaviour to future work.}
    \newtext{\item While CR and BWCR effectively prevent over-fitting to hard ground truth labels in ambiguous pixels, they fail (in most cases) to improve segmentation accuracy as compared to DA.}
    \item In the \emph{large} training set experiments for ACDC, CR and BWCR exhibit worse calibration than other methods. The segmentation accuracy is very high for all methods, but CR and BWCR still provide soft probabilities near boundaries thus causing poorer calibration.
    
\end{enumerate}

\input{6_exps_part4}
\input{6_exps_part5}
\input{6_exps_part6}

\clearpage

%% file: 6_exps_part4.tex
\begin{table}[!t]
        \begin{center}
        \begin{tabular}{c|ccc|ccc|ccc}

        \hline
        \multicolumn{10}{c}{\textbf{NCI}}\\

        \hline
         &
        \multicolumn{3}{c|}{$N_{\textrm{tr}} = 6$} & \multicolumn{3}{c|}{$N_{\textrm{tr}} = 12$} & \multicolumn{3}{c}{$N_{\textrm{tr}} = 36$} \\
        
        \cline{2-10}
        \multirow{-2}{*}{Method} &
        Dice $\uparrow$ & ECE $\downarrow$ & TACE $\downarrow$ &
        Dice $\uparrow$ & ECE $\downarrow$ & TACE $\downarrow$ &
        Dice $\uparrow$ & ECE $\downarrow$ & TACE $\downarrow$ \\

        \hline
        Baseline &
        55$\pm$16 & 41$\pm$16 & 15$\pm$4 & 
        58$\pm$17 & 39$\pm$16 & 16$\pm$5 & 
        68$\pm$13 & 27$\pm$11 & 13$\pm$4 \\ 

        DA \cite{zhang2020generalizing} &
        66$\pm$13 & 24$\pm$12 & 11$\pm$4 & 
        \textbf{69$\pm$13} & 23$\pm$11 & 12$\pm$4 & 
        \textbf{75$\pm$11} & 13$\pm$9 & 7$\pm$3 \\ 

        SVLS \cite{islam2021spatially} &
        66$\pm$14 & 23$\pm$13 & 11$\pm$4 & 
        68$\pm$14 & 20$\pm$11 & 9$\pm$3 & 
        75$\pm$12 & 14$\pm$9 & 7$\pm$2 \\ 

        MLS \cite{murugesan2022calibrating} &
        66$\pm$14 & 31$\pm$15 & 13$\pm$5 & 
        68$\pm$14 & 22$\pm$10 & 11$\pm$4 & 
        75$\pm$12 & 14$\pm$10 & 7$\pm$2 \\ 

        CR (Ours) &
        65$\pm$14 & 18$\pm$14 & 6$\pm$3 & 
        68$\pm$14 & 18$\pm$14 & 6$\pm$3 & 
        73$\pm$12 & 9$\pm$9 & 4$\pm$2 \\ 
        
        BWCR (Ours) &
        \textbf{67$\pm$13} & \textbf{14$\pm$12} & \textbf{5$\pm$3} & 
        \textbf{69$\pm$13} & \textbf{13$\pm$12} & \textbf{3$\pm$1} & 
        \textbf{75$\pm$11} & \textbf{7$\pm$7} & \textbf{3$\pm$1} \\ 

        \hline
        \hline
        
        \multicolumn{10}{c}{\textbf{ACDC}}\\

        \hline
         &
        \multicolumn{3}{c|}{$N_{\textrm{tr}} = 5$} & \multicolumn{3}{c|}{$N_{\textrm{tr}} = 10$} & \multicolumn{3}{c}{$N_{\textrm{tr}} = 95$} \\
        
        \cline{2-10}
        \multirow{-2}{*}{Method} &
        Dice $\uparrow$ & ECE $\downarrow$ & TACE $\downarrow$ &
        Dice $\uparrow$ & ECE $\downarrow$ & TACE $\downarrow$ &
        Dice $\uparrow$ & ECE $\downarrow$ & TACE $\downarrow$ \\

        \hline
        Baseline &
        58$\pm$15 & 37$\pm$17 & 18$\pm$6 & 
        67$\pm$15 & 22$\pm$13 & 8$\pm$6 & 
        86$\pm$6 & 8$\pm$6 & 6$\pm$2 \\ 

        DA \cite{zhang2020generalizing} &
        \textbf{76$\pm$12} & 20$\pm$14 & 10$\pm$4 & 
        82$\pm$8 & 11$\pm$9 & 6$\pm$3 & 
        \textbf{90$\pm$3} & 4$\pm$3 & \textbf{3$\pm$2} \\ 

        SVLS \cite{islam2021spatially} &
        75$\pm$12 & 19$\pm$14 & 8$\pm$4 & 
        \textbf{83$\pm$7} & 9$\pm$8 & 5$\pm$3 & 
        \textbf{90$\pm$3} & \textbf{3$\pm$3} & \textbf{3$\pm$2} \\ 

        MLS \cite{murugesan2022calibrating} &
        75$\pm$12 & 20$\pm$14 & 9$\pm$4 & 
        82$\pm$8 & 12$\pm$9 & 7$\pm$3 & 
        \textbf{90$\pm$3} & \textbf{3$\pm$3} & \textbf{3$\pm$2} \\ 

        CR (Ours) &
        75$\pm$12 & 13$\pm$12 & \textbf{5$\pm$2} & 
        81$\pm$8 & 8$\pm$6 & \textbf{4$\pm$1} & 
        88$\pm$4 & 7$\pm$2 & 4$\pm$1 \\ 
        
        BWCR (Ours) &
        75$\pm$11 & \textbf{11$\pm$10} & \textbf{5$\pm$2} & 
        82$\pm$8 & \textbf{8$\pm$5} & \textbf{4$\pm$1} & 
        89$\pm$3 & 8$\pm$2 & 4$\pm$1 \\ 
        \hline
        
        \end{tabular}

        \caption{Quantitative results reported as $\%$ average $\pm$ $\%$ standard deviation over test volumes and $3$ experiment runs.
        For brevity, TACE values are scaled by $10$.
        The best values in each column are highlighted, with the winner for tied averages decided by lower standard deviations.
        Paired permutation tests ($n = 10000$) show that ECE and TACE improvements of BWCR over all other methods are statistically significant with $p < 0.001$, for all except the ACDC \emph{large} training size setting.}
        \label{tab_quant_results}
        \end{center}
\end{table}

%% file: 6_exps_part5.tex
\begin{figure}[!t]
    \centering
    \includegraphics[width=\textwidth, trim={0cm 10.75cm 1cm 0cm},clip]{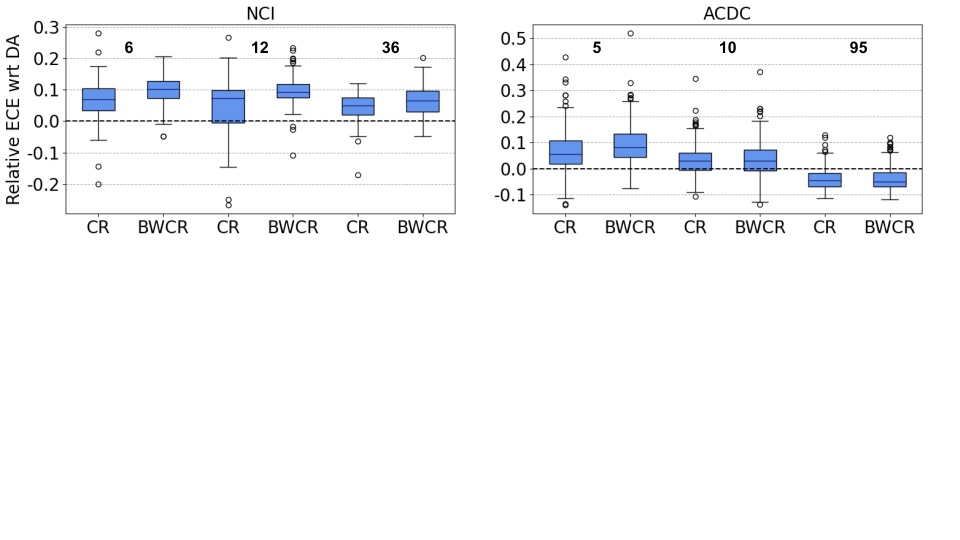}
    \caption{Subject-wise improvement in ECE due to
    \rephrasedtext{CR and BWCR, relative to DA.}
    Numbers between each set of CR and BWCR boxes indicate $N_{\textrm{tr}}$. Advantages of the proposed method are particularly prominent for small training set sizes.}
    \label{fig_subjectwise_ECE}
\end{figure}

%% file: 6_exps_part6.tex
\begin{figure}[!b]
    \centering
    \includegraphics[width=0.93\textwidth, trim={0cm 0cm 0cm 0cm},clip]{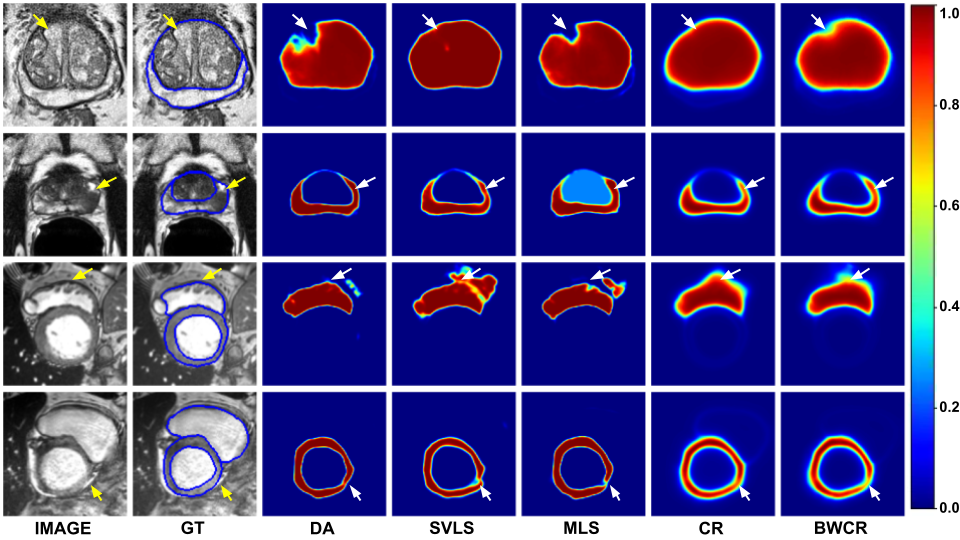}
    \caption{Qualitative comparison of calibration results for NCI CG (row 1), PZ (row 2), ACDC RV (row 3), and MY (row 4). Arrows point to spatially varying uncertainty predicted by the proposed method in ambiguous regions.}
    \label{fig_qualitative_results}
\end{figure}

%% file: 7_discuss.tex
\section{Conclusion}~\label{sec_discuss}
We developed a method for improving calibration of segmentation neural networks by noting that consistency regularization mitigates overfitting to ambiguous labels, and building on this understanding to emphasize this regularization in pixels most likely to face label noise.
Future work can \rephrasedtext{extend} this approach for lesion segmentation and / or $3$D models, \rephrasedtext{explore} the \rephrasedtext{effect} of other consistency loss functions \newtext{(e.g. cosine similarity or Jensen-Shannon divergence)}, \rephrasedtext{develop} other strategies to identify pixels that are more prone to ambiguity, or \rephrasedtext{study} the behaviour of improved calibration on out-of-distribution samples.

\newtext{\section*{Acknowledgements} This research is supported by NIH NIBIB NAC P41EB015902, IBM, and the Swiss National Science Foundation under project P500PT-206955.}

%% file: 8_bib.tex
%
%
%
\clearpage
\newpage
\bibliographystyle{splncs04}
\bibliography{miccai23}